\documentclass[letterpaper,10pt,conference]{ieeeconf} 
\IEEEoverridecommandlockouts  
\usepackage{graphicx}
\usepackage{url}
\usepackage{xcolor}
\usepackage{subfig}
\usepackage{hyperref}
\usepackage{amsmath}

\begin{document}

\title{Magnetic Field Sensing for Pedestrian and Robot  Indoor Positioning}

\author{Leonid Antsfeld and Boris Chidlovskii\\
Naver Labs Europe, Meylan 38240, France \\
firstname.lastname@naverlabs.com}




\maketitle
\begin{abstract}
In this paper we address the problem of indoor localization using magnetic field data in two setups, when data is collected by (i) human-held mobile phone and (ii) by localization robots that perturb magnetic data with their own electromagnetic field. 
For the first setup, we revise the state of the art approaches and propose a novel extended pipeline to benefit from the presence of magnetic anomalies in indoor environment created by different ferromagnetic objects. 
We capture changes of the Earth's magnetic field due to indoor magnetic anomalies and transform them in multi-variate times series. 
We then convert temporal patterns into visual ones. We use methods of Recurrence Plots, Gramian Angular Fields and Markov Transition Fields to represent magnetic field time series as image sequences. We regress the continuous values of user position in a deep neural network that combines convolutional and recurrent layers.
For the second setup, we analyse how magnetic field data get perturbed by robots' electromagnetic field. We add an alignment step to the main pipeline, in order to compensate the mismatch between train and test sets obtained by different robots. We test our methods on two public (MagPie~\cite{magpie_article} and IPIN'20~\cite{potorti20ipin}) and one proprietary (Hyundai department store) datasets.
We report evaluation results and show that our methods outperform the state of the art methods by a large margin. 
\end{abstract}
\maketitle

\section{Introduction}
\label{sec:introduction}

Ubiquitous location-based services have recently attracted a great deal of attention. They require reliable positioning and tracking technology for mobile devices that works outdoors as well as indoors. While navigation satellite systems such as GPS (Global Positioning System) already provide reliable positioning outdoors, a corresponding solution is yet to be found for an indoor environment where GPS signals cannot penetrate and provide sufficient accuracy. Over the past few years, indoor positioning and localization have become an area for intensive research and development, which shows the importance of this area~\cite{zafari2017a}.


One important category of localization systems is {\it infrastructure free}~\cite{gu_indoor_2019}. This category includes positioning systems based on inertial measurement unit (IMU) and a magnetometer. These sensors have an additional advantage of a low cost over WiFi or Bluetooth sensors commonly used for localization.  
One known infrastructure-free example is a Pedestrian Dead Reckoning (PDR) system that utilizes the accelerometer and gyroscope of the smartphone to track the user's path; it however provides a relative position only and always needs a starting position. The magnetic field based approach also has a good potential to produce good results, because elements containing iron elements (walls, pillars, windows) often create unique magnetic anomalies~\cite{magpie_article,kalyan_16_locateme}. Recent works on localization using a magnetic field witness the growing interest to this research area and its potential to contribute to the localization problem in general~\cite{ashraf20sensors,ashraf_floor_2019,ashraf20minloc,amid_article}.

Indoor magnetic signatures are disturbances of the Earth's magnetic field induced by various ferromagnetic objects, such as walls, pillars, doors, elevators, etc. 
These anomalies become dominant at small distances from such objects. 
A large number of indoor ferromagnetic objects and the disturbances they induce form signatures with unique patterns; this allows to classify these signatures based on their patterns.

\begin{figure}[ht]
\centering
\includegraphics[width=0.9\columnwidth]{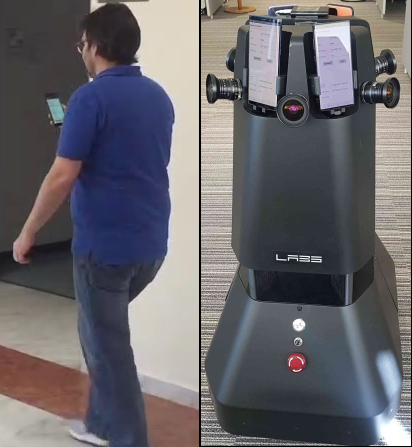}
\quad 
\caption{Left) human-held mobile phone, right) Naver Labs navigation robot, with two mobile phones installed on the front side and two ones on the back side.}
\label{fig:photos}
\end{figure}

In the human-held mobile phone setup (see Figure~\ref{fig:photos}.left), we build on the state-of-the-art AMID method~\cite{amid_article} which identifies {\it landmarks} in the magnetic map and processes the localization task as landmark classification. 

We revise and extend~\cite{amid_article} to benefit from the recent advances in convolutional and recurrent deep networks. First, we exploit the sequential nature of magnetic data collected when users navigate indoor thus forming multi-variate time series. Second, we propose new methods to convert the multi-variate time series into visual representations. These images aim to capture different magnetic patterns similarly to using multiple cameras in video streams~\cite{khoshro15}. They form multi-channel input to convolutional layers that extract {\it position vector embeddings} which, third, we feed to fully connected (FC) layers. The FC layers can be trained in classification mode to predict the closest landmark~\cite{amid_article}, or in regression mode to directly estimate the user's position coordinates.

Evaluations show that both approaches, the regression-based and landmark-based,
fail in very similar situations. These failures are caused by the similar magnetic signatures the system faces in very different places. Indeed, the same magnetic anomalies are caused by the same ferromagnetic objects, for example, identical armature pillars placed in the different corners of a hall.

To help the system disambiguate similar magnetic patterns, we capture the {\it localization context} with the recurrent layers which replace the fully connected layers in the multi-channel deep regression.
The convolutional layers extract position embeddings while recurrent layers encode the localization context in internal states to disambiguate similar patterns, in the same way as RNNs use context to disambiguate word understanding in NLP tasks~\cite{yin17nlp}. 

Like other models for sequential data, our models are exposed to the bootstrapping problem; they need to accumulate magnetic sensor data to produce a first prediction, in practice such an accumulation can take $7-10$ seconds. 
The recurrent models have another constraint: they require knowing the starting point of a trial. Our solution is to start with an approximated location from the landmark-based classification or CNN-based regression models, or from other sensors, such as Wi-Fi signals. All of them being a subject of noisy predictions, we test our system under the starting point estimation error.

Intensive evaluations on the MagPIE dataset~\cite{magpie_article} and IPIN'20 dataset~\cite{potorti20ipin} show that all these improvements contribute to the robust and accurate localization pipeline.
Indeed, we are able to reduce the localization error for three MagPie dataset buildings to $0.30 - 1.05$ m, thus improving by the large margin the baseline method with $0.95-4.49$ m error.


Despite these good results in the {\it human-held mobile phone setup}, its na\"ive transfer to the {\it robot-based setup} fails. Indeed, when robots are used to collect magnetic field data (Figure~\ref{fig:photos}.right), Earth's magnetic field is disturbed, in addition to nearby ferromagnetic objects, by electromagnetic field generated by robot's engines, batteries, wires etc.   
As result, magnetic data collected by a robot 
differs in both values and scale from data collected by a user walking with a smartphone or by another robot. 
To make the collected magnetic data usable for localization tasks, we introduce an {\it auxiliary alignment step} before pushing data in the localization pipeline. 
We propose two alignment methods and successfully test them on Hyundai department store dataset. 


Main contributions of this work are the following:
\begin{enumerate}
   \item We address the problem of magnetic field based indoor localization and propose a unique pipeline for two setups, when data is collected by human-held mobile phones and by localization robots. 
   \item In the first setup, we replace the landmark-based classification~\cite{amid_article} 
    with the deep regression. We convert the multi-variate time series into multi-channel 2D image sequences; replacing pattern detection in time series by pattern detection in images aim to benefit from the recent progress in convolutional and recurrent neural networks.
    
    \item In the second setup, we complete the localization pipeline with an additional alignment step aimed to compensate the disturbances caused by robots' electromagnetic field. 
    
    
    \item Evaluation on three magnetic field (MF) datasets show that our methods report a low localization error and outperform the state of the art methods by a large margin. It makes the MF-based positioning competitive and comparable to Wi-Fi, Bluetooth and PDR methods, however requiring no infrastructure investment.
    
\end{enumerate}

\section{Related Work}
\label{sec:related_works}

With the wide expansion of modern smartphones with embedded sensors, many indoor localization solutions have emerged, including those based on user's activity recognition and navigation~\cite{song14}. Similarly, magnetic ﬁeld-based indoor positioning systems rely on the use of a smartphone built-in magnetic sensor~\cite{ashraf20sensors,ashraf_floor_2019,ashraf20minloc,amid_article,song14}.

The first geomagnetic ﬁeld-based indoor positioning systems use the K-Nearest Neighbor (KNN) algorithm to estimate the user’s position~\cite{chung11,song14}. Such a system gets a number of position candidates obtained through matching the user
magnetic signature against the magnetic database and uses these candidates to predict the user’s current location. 

LocateMe~\cite{kalyan_16_locateme} was the first indoor localization system using magnetic sensors data only. It processed indoor magnetic signatures as a combination of the Earth's magnetic field and the fields generated by ferromagnetic objects. 
The impact of these structures becomes dominant as the distance to such an object decreases. Consequently, these signatures are displaying a uniqueness in their patterns, that allows classifying signatures based on their patterns.


AMID~\cite{amid_article} was the first indoor positioning system that recognizes magnetic sequence patterns using a deep neural network (DNN). Features extracted from magnetic data sequences are fed to a DNN to classify the sequences by patterns that are generated by nearby magnetic landmarks. 

It first used the Recurrence Plots (RP) from the time series analysis to categorize the sequence patterns. 
As RPs can not distinct the sequence’s direction, additional features of trend, sequence length, and peak values are extracted to complete the input to DNN. Sequence length and values of reference points are used to classify the monotonic shape of magnetic sequences. Most of the landmark candidates have much higher or lower values than the mean magnetic intensity. Candidates of low interest are filtered out using a threshold. 
The DNN in AMID consists of a convolution neural network (CNN) for analyzing image features and a multi-layer perceptron (MLP) for magnetic landmarks classification. Every RP is converted into a $32 \times 32$ image as an input to the CNN. The locations are estimated from the locations of predicted landmarks.
 

\section{Magnetic sensors data}
\label{sec:magneticdata}

The goal of any smartphone-based localization system is to determine the user's position by analysing the smartphone's sensors data. As the floor of a building is often accurately detected from the {\it pressure sensor} data, we can reduce 3D localization to a simpler, 2D localization problem, where the user's position is described by two values, $pos=(pos_x, pos_y)$.

\begin{figure}[ht]
\centering
\subfloat[local reference frame used by Android]{\includegraphics[width = 0.3\columnwidth]{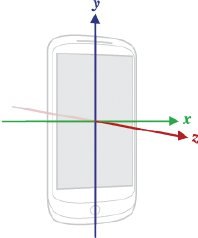}} 
\subfloat[global reference frame]{\includegraphics[width = 0.3\columnwidth]{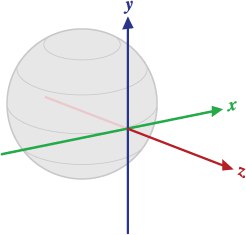}} 
\subfloat[smartphone rotation angles 
]{\includegraphics[width = 0.3\columnwidth]{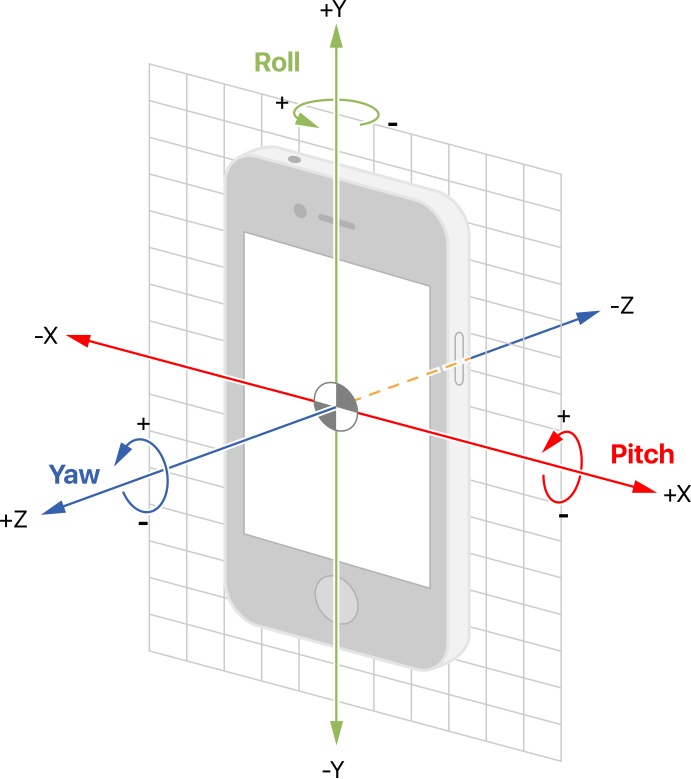}}
\caption{Smartphone reference frames and rotation angles}
\label{fig:smartphone_rotations}
\end{figure}

In a smartphone,
accelerometer, gyroscope and magnetometer report their readings in the local reference frame. 
Accelerometer and gyroscope data can be used to convert magnetic sensor values from local to global reference frame (see Figure~\ref{fig:smartphone_rotations}). 
In the following we assume that the MF data 
refers to the global reference frame and forms a multivariate time series composed of three values $m=(m_x, m_y, m_z)$ at timestamp $t$.

Both train and test datasets include multiple trials. Any train trial includes MF data and ground truth positions, $D_{train} = (t, m, pos)$, where $t$ is the measurement timestamp. In a testing trial, ground truth positions are unavailable, $D_{test} = (t, m)$.

Mobile phone sensors are generally 
captured at different timestamps and different frequencies.
In the following, we assume that the sensor readings are synchronised and aligned at the same timestamps.

\paragraph{Magnetometer calibration}
Build-in magnetometers are low cost sensors and their MF measurements are often corrupted by errors including sensor fabrication issues and the magnetic deviations induced by the smartphone platform. Therefore a proper calibration of the magnetometer is critical to achieve high accuracy~\cite{yu21calibration}.

The phone orientation in the space is described 
by rotation angles (yaw, pitch, roll), as shown in  Figure~\ref{fig:smartphone_rotations}. 
These rotation angles can be obtained directly from the phone or reconstructed from the IMU sensors~\cite{heading_estimation}.

\begin{figure}[ht]
\centering
\subfloat{\includegraphics[width =0.9 \columnwidth]{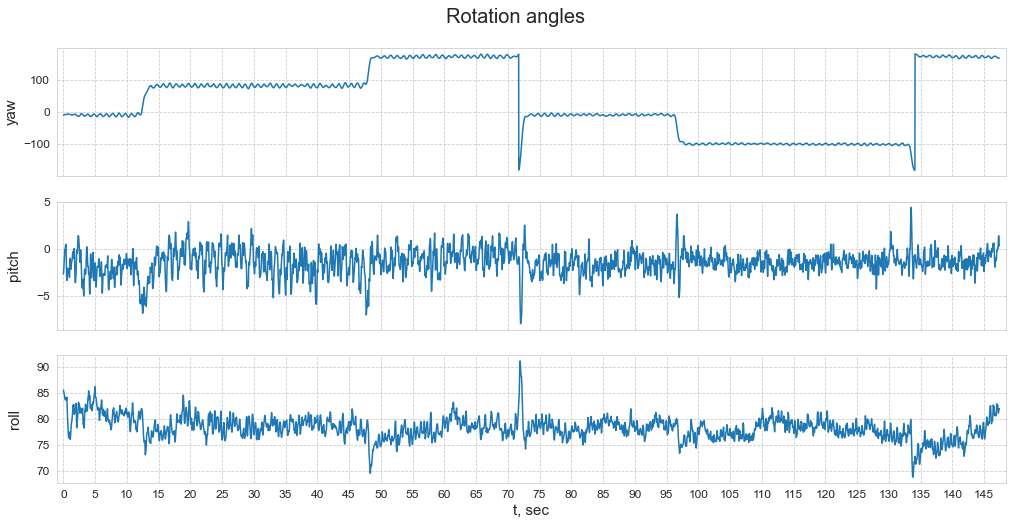}}
\caption{Rotation angles for one trial.}
\label{fig:trial_rotations}
\end{figure}

While walking, the user can hold the phone in various positions (upward in front of him/her, near the ear, in a pocket, etc.). Even during a short period of time, the position of the phone can change significantly (see example in Figure~\ref{fig:trial_rotations}). As phone sensors generate their measurements in the local reference frame, we constantly use rotation angles to convert them in the global reference frame.
\section{Deep neural network}
\label{sec:deep}
The core of MF-based localization is implemented as a deep neural network (DNN). DNN learning process consists of the objective function optimization and multiple network updates though the gradient back-propagation. All our models follow either classification or regression  approach. 
Classification models are trained with the Cross Entropy (CE) loss. Regression models are trained to minimize the Mean Square Error (MSE), Mean Absolute Error (MAE) or Huber loss. 

\subsection{From time series to image representation}
\label{ssec:1d-2d}

In a single trial, MF values form a multivariate time series where each observation consists of three values $m=(m_x, m_y, m_z)$ along axes $x$, $y$ and $z$. The orthogonality of the three axes is not obligatory. Moreover, some orientations appear to be more important than others. We test alternative combinations and projections, such as $m_{xy} = \sqrt{m_x^2 + m_y^2}$ and $m_{xyz} = \sqrt{m_x^2 + m_y^2 + m_z^2}$. The choice of optimal projections of MF values is a model hyper-parameter. In experiments, using $m_{xyz}$ projection improves the performance in majority of cases. 

After projections, the MF values are fed to DNN. Consider the feature generation step for a 1D time series $V^{'}= \{v_1, v_2, \ldots,v_n\}$. The sliding window technique cuts the time series into a sequence of overlapping segments of length $m \ll n$. Optimal values for windows size are around $7-10$ seconds, for the windows step of $0.2-1$ seconds. For every segment we apply a nonlinear transformation from 1D time series to 2D images thus transforming MF patterns into visual ones. AMID~\cite{amid_article} used convolutional neural networks to analyze time series segments represented as recurrence plots (RPs). As RPs address one specific type of recurrence in time series, we consider alternative methods for encoding 1D time series into 2D visual patterns, in particular Gramian Angular Summation/Difference Fields (GASF/GADF) and Markov Transition Fields (MTF)~\cite{gaf_mtf_article}.
We describe all these 
transformations below.

\paragraph{Recurrence plots}
Recurrence plots (RPs) have been widely used for time series analysis in various applications~\cite{hirata21reccurence,surveyRP2020}.
For a time series segment $V=\{v_1,v_2,\ldots,v_m\}$, RP is calculated for the Euclidean metric as follow:
\begin{equation}
\begin{tabular}{c}
$d_{ij} = || v_i - v_j || \quad \forall i, j \in 1..m$, \\
$RP_{ij} = 1 - \frac{d_{ij}}{\max (d)}$.
\end{tabular}
\label{eq:pr}
\end{equation}

The method can be extended to any pairwise distance metric. 
We tested twelve metrics available in Python's scipy package and detected {\it canberra} distance as performing the best over all evaluation settings. 

\paragraph{Gramian Angular Fields}
In the Gramian Angular Field (GAF), a time series segment $V$ is represented in a polar coordinate system. 
Values in the segment are re-scaled so that all values fall in the interval $[-1, 1]$ in order to be represented as polar coordinates achieved by applying angular cosine,
\begin{equation}
\begin{tabular}{c}
$\tilde{v_i} = \frac{(v_i - \max (V)) + (v_i - \min (V))}{\max (V) - \min (V)}$, \\
$\theta_i = \arccos (\tilde{v_i}), \ -1 \le \tilde{v_i} \le 1, \  \tilde{v_i} \in \tilde{V}.$
\end{tabular}
\label{eq:gaf}
\end{equation}
The polar-encoded segment of length $m$ is then transformed into a $m \times m$ matrix. We include in our pipeline two variants of Gramian Angular Field, Gramian Angular Summation Field (GASF) and Gramian Angular Difference Field (GADF), defined as follows
\begin{equation}
\begin{tabular}{c}
 $GASF_{ij} = \cos (\theta_i + \theta_j)$, \\ 
 $GADF_{ij} = \sin (\theta_i - \theta_j)$.
 \end{tabular}
 \label{eq:gaf2}
\end{equation}

\paragraph{Markov Transition Field}
The Markov Transition Field (MTF) considers time series as a outcome of Markov process. The method builds the Markov matrix of quantile bins after discretization and encodes the dynamic transition probability in a quasi-Gramian matrix~\cite{gaf_mtf_article}.

For a time series segment $V$, we identify its $Q$ quantile bins and assign each $v_i$ to the corresponding bins $q_j, j \in [1, Q]$. Thus we construct a $Q \times Q$ weighted adjacency matrix $W$ by counting transitions among quantile bins in the manner of a first-order Markov chain along the time axis, as follows:
\begin{equation*}
MTF = 
\begin{bmatrix}
w_{ij|v_1 \in q_i, v_1 \in q_j} & \cdots & w_{ij|v_1 \in q_i, v_n \in q_j} \\
w_{ij|v_2 \in q_i, v_1 \in q_j} & \cdots & w_{ij|v_2 \in q_i, v_n \in q_j} \\
\vdots & \ddots & \vdots \\
w_{ij|v_n \in q_i, v_1 \in q_j} & \cdots & w_{ij|v_n \in q_i, v_n \in q_j}
\end{bmatrix}
\end{equation*}

\begin{figure}[h]
\centering
\subfloat{\includegraphics[width=0.9\columnwidth]{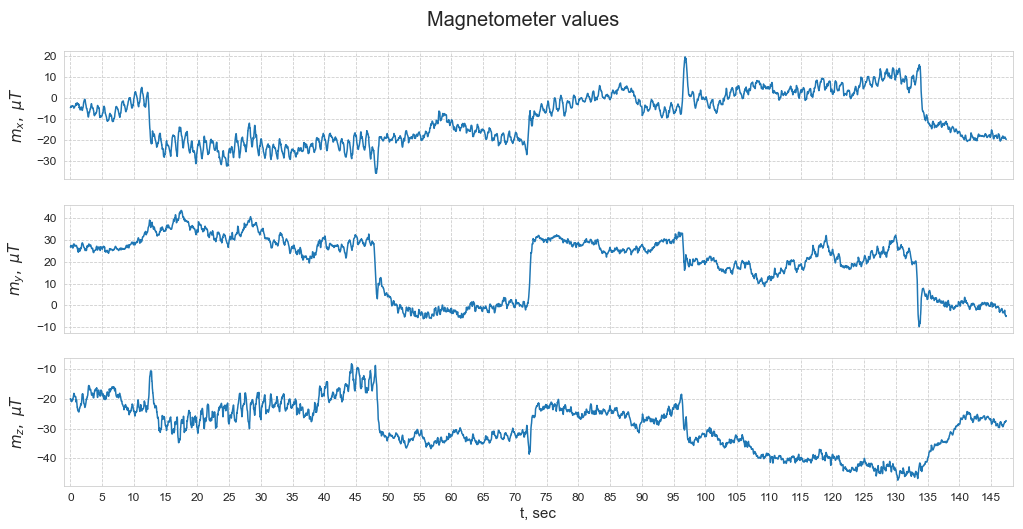}}
\caption{Three magnetometer values for one trial.}
\label{fig:trial_magnetometer_values}
\end{figure}

We visualize all 1D-to-2D transformations 
for one trial from the MagPIE dataset~\cite{magpie_article} with the window size of 7 s, step of 1 s, image size 100 and {\it canberra} distance. Figure~\ref{fig:trial_magnetometer_values} shows the MF time series 
for the selected trial; this trial was also used for angle rotations in Figure~\ref{fig:trial_rotations}.

\begin{figure}[ht]
\centering
\subfloat[Sequence of ten Recurrence plots]
{\includegraphics[width = \columnwidth]{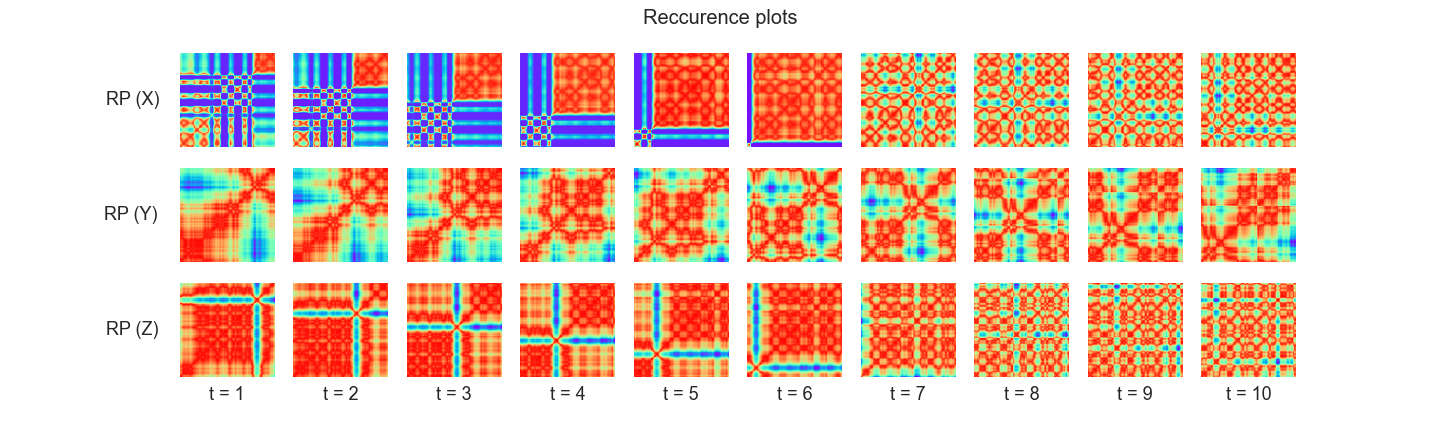}}
\hfill
\subfloat[Consecutive Gramian Angular Summation Field plots]
{\includegraphics[width = \columnwidth]{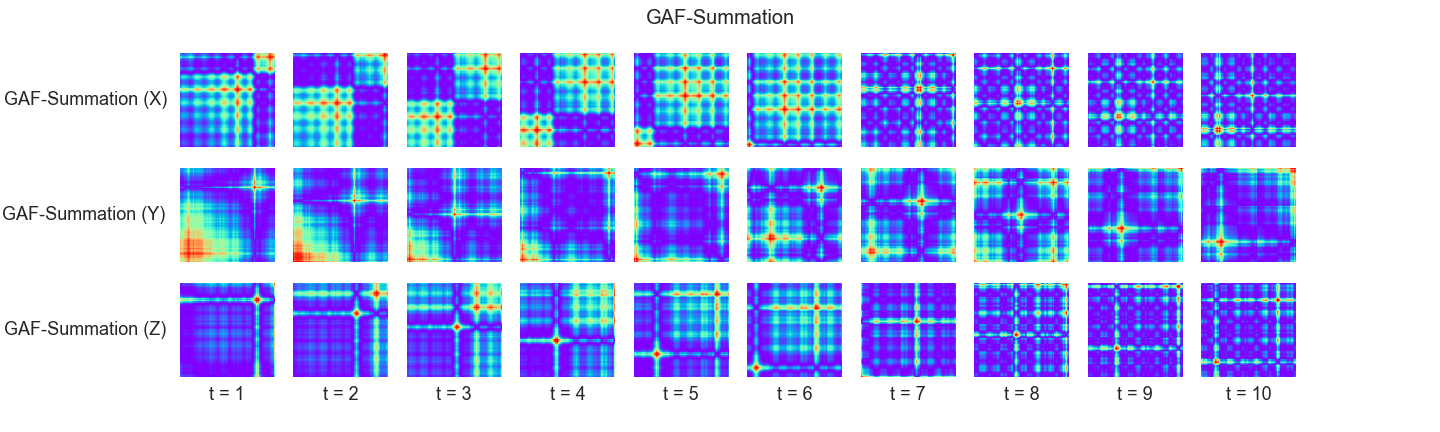}}
\hfill
\subfloat[10 consecutive Gramian Angular Difference Field plots]
{\includegraphics[width = \columnwidth]{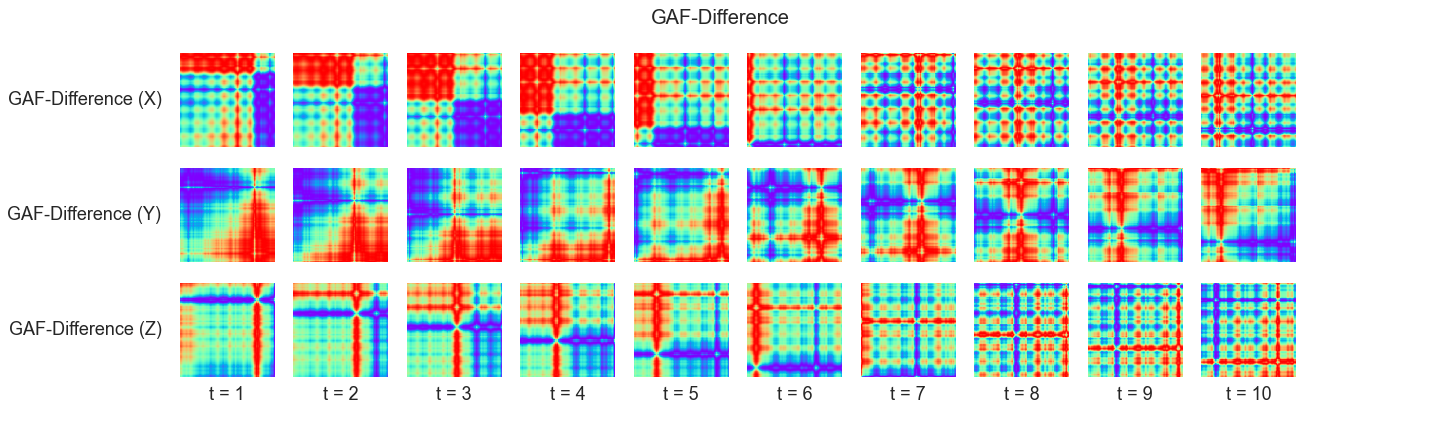}}
\hfill
\subfloat[10 consecutive Markov Transition Field plots]
{\includegraphics[width = \columnwidth]{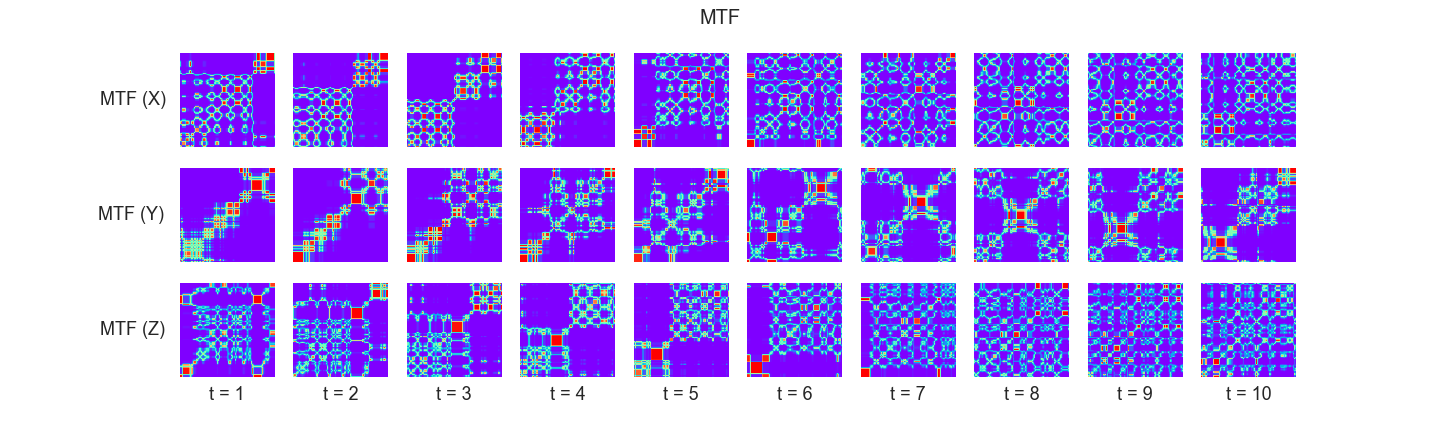}}
\caption{Visualization for four 1D-to-2D transformation methods: a) RP, b) GASF, c) GADF, d) MTF.} 
\label{fig:images}
\end{figure}

Figures~\ref{fig:images}.a,b,c,d show the results of transforming the time series in image sequences using RP, GASF, GADF and MTF methods presented above. 
Images are generated for three values $(m_x, m_y, m_z)$ separately.

\subsection{Landmark identification and deep classification}
\label{chap:landmark_identification}

We start by extending the AMID~\cite{amid_article} baseline which constructs magnetic maps, detects magnetic landmarks and trains a model to classify a correct position in one of the landmarks. 
Landmark detection consists of the three steps described below:

\begin{itemize}
    \item \textit{Local Minima/Maxima Detection:} Magnetic landmarks are ferromagnetic objects having larger or smaller magnetic intensities than their surroundings. Therefore, magnetic landmark candidates can be identified by finding the local minima/maxima in a magnetic map.
    \item \textit{Magnetic Landmark Candidate Refinement:} Not all local minima/maxima points can be used as magnetic landmarks. Multiple outliers may exist, this depends on the indoor environment and magnetic landmark characteristics. In some areas, magnetic intensity rarely changes; in other magnetic landmarks, magnetic intensities fluctuate over time. 
    Such fluctuations generate clusters of local minima/maxima. Magnetic landmark candidate refinement helps to solve the problem. AMID~\cite{amid_article} uses a distance-based hierarchical tree is used to group these points as one magnetic landmark candidate.
    \item \textit{Magnetic Landmark Selection:} Most of the landmark candidates have much higher or lower values than the mean magnetic intensity. However, the magnetic intensity of some candidates is too close to the average intensities. To filter these candidates out, manually selected thresholds are applied.
\end{itemize}

Once the magnetic landmarks are selected, the localization task is reduced to identifying the closest landmark and therefore can be solved as a classification problem.

\begin{figure}[ht]
\centering
\subfloat{\includegraphics[width =\columnwidth]{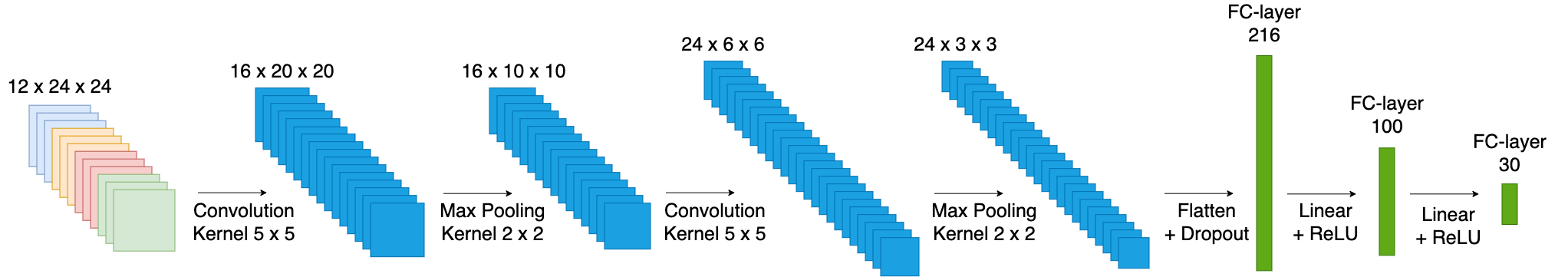}}
\caption{$N$-channel CNN+FN landmark classification network.}
\label{fig:class_model}
\end{figure}

Our first extension of the AMID baseline is to replace one RP channel with $N$ channels, where $N$=12 include 3 channels for each of RP, GASF, GADF, and MTF. The classification architecture consists of two convolutional layers that extract position embeddings, and two fully connected layers for classification
(Figure~\ref{fig:class_model}). 
Loss function is the cross-entropy loss.

\subsection{Deep Regression}
\label{ssec:deep_regression}

Experiments with the extended landmark-based classification model shows that its performance depends on the quality of magnetic map, a good coverage of the indoor space and a manual tuning of the critical pre-processing thresholds which are not a part of the training process. In the following we replace the landmark-based classification with direct user's position regression.

The first, {\it CNN+FN regression model} replicates the landmark classification  architecture 
but the output layer has only 2 variables, $pos_x$ and $pos_y$. It uses MSE, MAE or Huber loss as the objective minimization function.

The CNN+FN regression makes the system independent of the MF map quality and the selected landmarks. But evaluations show that both landmark-based an CNN-based regression fail when facing similar magnetic patterns in different locations. To help disambiguate them we take the navigation context into account and replace FN layers with recurrent layers.
Recurrent neural networks (RNNs) are widely used when working with sequential, regular timestamp-based data. However, they require to change the training protocol and process data trial-by-trail and not point-by-point manner. For each track, the position estimations are generated sequentially, where the previous estimations $(pos_x, pos_y)_{1..t}$ are used to predict the next position, $(pos_x, pos_y)_{t+1}$.

\begin{figure}[ht]
\centering
\subfloat{\includegraphics[width = \columnwidth]{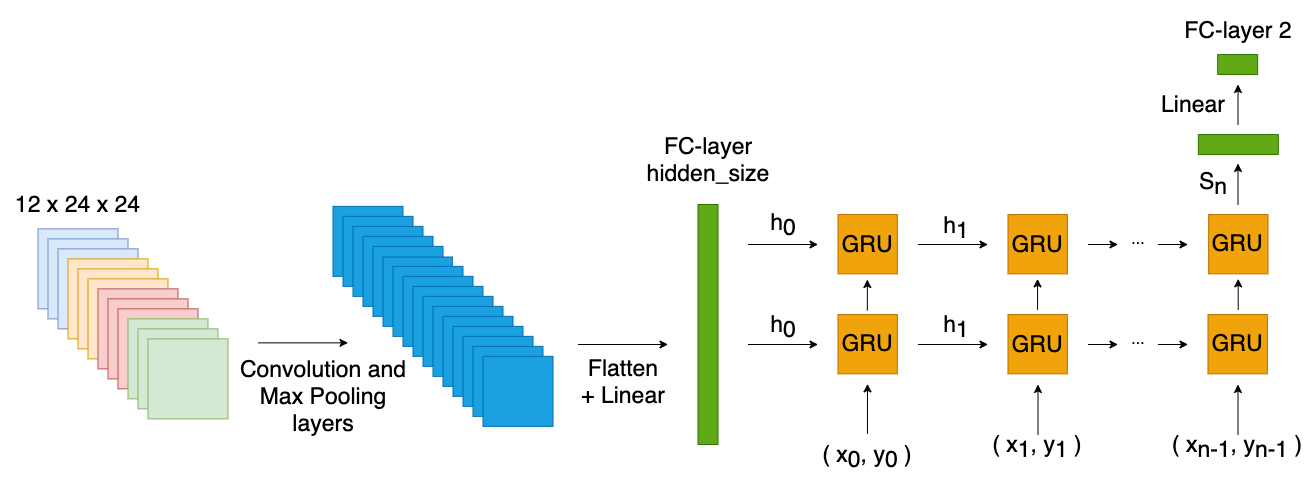}}
\caption{$N$-channel CNN+RNN deep regression network.}
\label{fig:rnn_model}
\end{figure}

The {\it multi-channel CNN+RNN deep regression} preserves convolutional layers to extract position embeddings but replaces FC layers with recurrent layers. 
Figure~\ref{fig:rnn_model} shows the architecture which completes CNNs with 2-layer one-directional RNN on GRU cells~\cite{gru_article}. During evaluation phase, we also tested other configurations, such as 3 and more layers, bi-directional layers, but a small improvement was obtained for the price of more computationally intensive training.

To bootstrap, we make RNNs accept input sequences of varying lengths. At the start where only the first point is available, it uses only this point to make the prediction. Then it predicts the third point based on 2 previous points, etc. thus increasing the window length to a predefined maximum size. As a result, the system can use the sequence of all the previous points. As evaluations show, the best results are obtained with the maximum window size of 10-20 points. 

There is another issue in RNN-based localization, namely, the approach assumes knowing the first ground truth point. In real life, this information is usually unavailable. Instead, alternative sensors and localization components can be used to get the first point position, for example from Wi-Fi or Bluetooth signals. As the average distance error of Wi-Fi-based localization~\cite{chidlovskii19} is 2-3 meters, we simulate this error in both train and test phase, by adding a random noise to the starting point position.

Training recurrent neural networks is a time- and resource-consuming process. To speed up RNN training, 
we reused some techniques developed for NLP tasks. In particular, we used the {\it teacher forcing} which, at the training phase, replaces the prediction by the ground truth with a probability $p_{teach}$.

\section{Robot magnetic field and localization}
\label{sec:aligment}

The localization pipeline presented in Sections~
\ref{sec:magneticdata} and~\ref{sec:deep} addresses the human-held mobile phone setup. In this section, we address some challenges on the robot-based setup.

\begin{figure}[ht]
\centering
{\includegraphics[width=0.49\columnwidth]{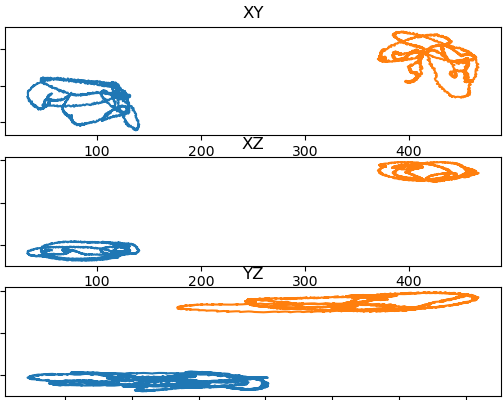}
\includegraphics[width=0.49\columnwidth]{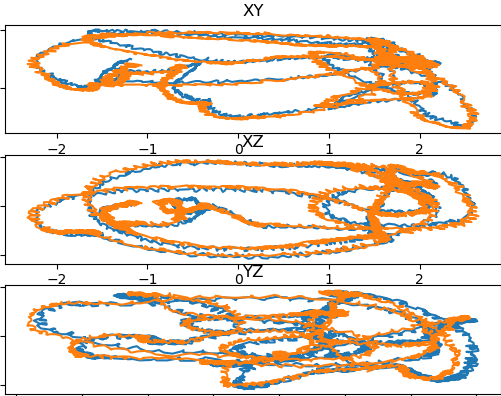}}
\caption{Example of magnetic data projections of train and test sets at the same track; a) before alignment and b) after alignment.}
\label{fig:align}
\end{figure}

Figure~\ref{fig:align} illustrates the mismatch caused by difference in two robot magnetic footprints we observed in Hyundai store dataset. Figure~\ref{fig:align}.a plots the projections of train set $D_1$ collected by robot $R_1$ (in \textcolor{orange}{orange}) and test set $D_2$ collected by robot $R_2$ (in \textcolor{blue}{blue}) in the {\it same} corridor, projected in 2D dimensions $x$-$y$, $x$-$z$ and $y$-$z$.
This mismatch has dramatic consequences; every localization model trained on $R_1$ data totally fails on $R_2$ data (see Table~\ref{tab:summary-hyundai} in Section~\ref{ssec:ipin_eval}). 
We therefore introduce an additional alignment step to compensate the mismatch. Figure~\ref{fig:align}.b shows the result of successful alignment after applying one of the methods presented in the following section. Once aligned, $R_1$ data and aligned $R_2$ data represent very similar 3D curves of nearby magnetic anomalies; this allows $R_1$ model to work well on aligned $R_2$ data, despite a multitude of small differences in their physical trajectories in the same corridor.

We use a common segment in the train and test tracks. Since 
the magnetic data along this track is expected to be very similar, we us the common segment to find an alignment transformation from the test set to the train set. 

We adopt the following alignment principle.
Assume that ``clean" MF data (collected by a human-hand mobile phone) is are represented as $(m, pos)$ where $m$ is a 3D vector $(m_x,m_y,m_z)$ am $pos=(pos_x, pos_y)$.
We then represent MF data collected by $R_1$ and $R_2$ as $(f_1(m_1),pos_1)$ and $(f_2(m_2),pos_2)$, respectively, where $f_1$ and $f_2$ are perturbation functions that the electromagnetic field of $R_1$ and $R_2$ exercises on vectors $m$, respectively. In other words, $f_i$ is an individual magnetic footprint of robot $R_i$, $i=1,2$.

We are now looking for a transformation of $f_2(m)$ values into $f_1(m)$ space, without knowing the exact value of ``clean" vector $m$. We form an alignment dataset $D_A$ as a set of pairs
$(f_1(m_1), f_2(m_2)), (m_1, pos_1) \in D_1, (m_2, pos_2) \in D_2$, such that $pos_1 = knnUnique(pos_2,k)$ and $dist(pos_1,pos_2)<\epsilon$. Here, $knnUnique(\cdot)$ function guarantees that $pos_1$ is among $k$-neighbours of $pos_2$ and, moreover, paired with $pos_2$ only. In other words, we form pairs of MF data items collected by two robots in the nearest, close and unique neighbourhood.

To populate the alignment dataset $D_A$ for Hyundai store, we set the distance threshold $\epsilon=$0.5 m. The alignment step requires identifying a common segment
in $D_1$ and $D_2$ data sets. We therefore split $R_2$ data in {\it alignment} and {\it true test} parts. The alignment part (about 5\%) refers a common segment (a corridor of 10 m long) navigated by both robots in the same direction.\footnote{Our pipeline includes data augmentation component aimed at increasing the dataset by synthesizing new tracks in direction opposite to the real ones. But for the sake of simplicity, the alignment step assumes the same direction.} Remaining 95\% of $R_2$ dataset is used for testing the $R_1$ model on aligned $R_2$ data. 

\subsection{Data Alignment} 
\label{ssec:opt_method}
We use the alignment dataset $D_A$ to find
an alignment function $g$ between the robot perturbation functions $f_1$ and $f_2$. A good alignment allows a model trained on train data $(f_1(m_1),pos_1)$ from $R_1$, to be applied to the aligned version of $R_2$ data, $(g(f_2(m_2)),pos_2)$.

Below we propose two approaches to find an optimal transformation $g$. Formally, we are looking for a optimal solution to the following optimization problem:
\begin{equation}
    g^* = argmin_{g \in {\cal G}} \sum_{(m_1,m_2)\in D_A} ||g(f_1(m_1)) - f_2(m_2)||_2,
    \label{eq:align}
\end{equation}
where $\cal G$ is a class of transformation functions, ${\cal G}: R^3 \rightarrow R^3$. Below we consider two options to solve the problem stated in Eq.(\ref{eq:align}).
\begin{enumerate}
\item 
We first try the class ${\cal G}_l$ of {\it linear transformations}, ${\cal G}_l=SO(3)$.
We model $g \in SO(3)$ as a $3\times 3$ matrix corresponding to rotating a 3D vector $f_1(m_1)$ into vector $f_2(m_2)$, for all vector pairs in dataset $D_A$. The optimal solution $g^*$ can be obtained using any standard linear optimization package (we used Python {\it numpy} package).

\item 
The class ${\cal G}_l$ of linear transformations appears to be rather limited. Indeed, a linear transformation makes an assumption that robot footprint functions $f_i$ depend on MF value $m$ only. In reality, they often depend on the robot's position ($pos$) and nearby magnetic anomalies. 

Our second attempt is therefore to consider the class ${\cal G}_n$ of {\it nonlinear deep transformations}. We model $g \in {\cal G}_n$ as a deep network that consists of 3 three FC layers. The network is trained to minimize the loss in Eq.\ref{eq:align} using pairs from the alignment dataset $D_A$. For the common segment in the Hyundai store set (10 m long corridor), $D_A$ includes ~2000 pairs. We train the network for 5000 epochs using the SGD optimizer with a cyclic learning rate scheduler \cite{cyclic_lr} (initial learning rate $1e^{-4}$, maximum learning rate $1e^{-3}$, step size $100$), the batch size is 32. In Section~\ref{sec:experiments} we test both alignment solutions. While the linear solution can work with a small alignment dataset, the deep solution with a large dataset $D_A$ requires a hyper-parameter fine-tuning.
\end{enumerate}

\section{Evaluation results}
\label{sec:experiments}

We evaluate our models on two public datasets, MagPIE~\cite{magpie_article} and IPIN'20~\cite{potorti20ipin}, and one proprietary dataset collected with navigation robots in the Hyundai Department Store shopping mall in Seoul, Korea.

\begin{figure*}[ht]
\centering
\subfloat[Loomis First Floor]{\includegraphics[width=0.65\columnwidth]{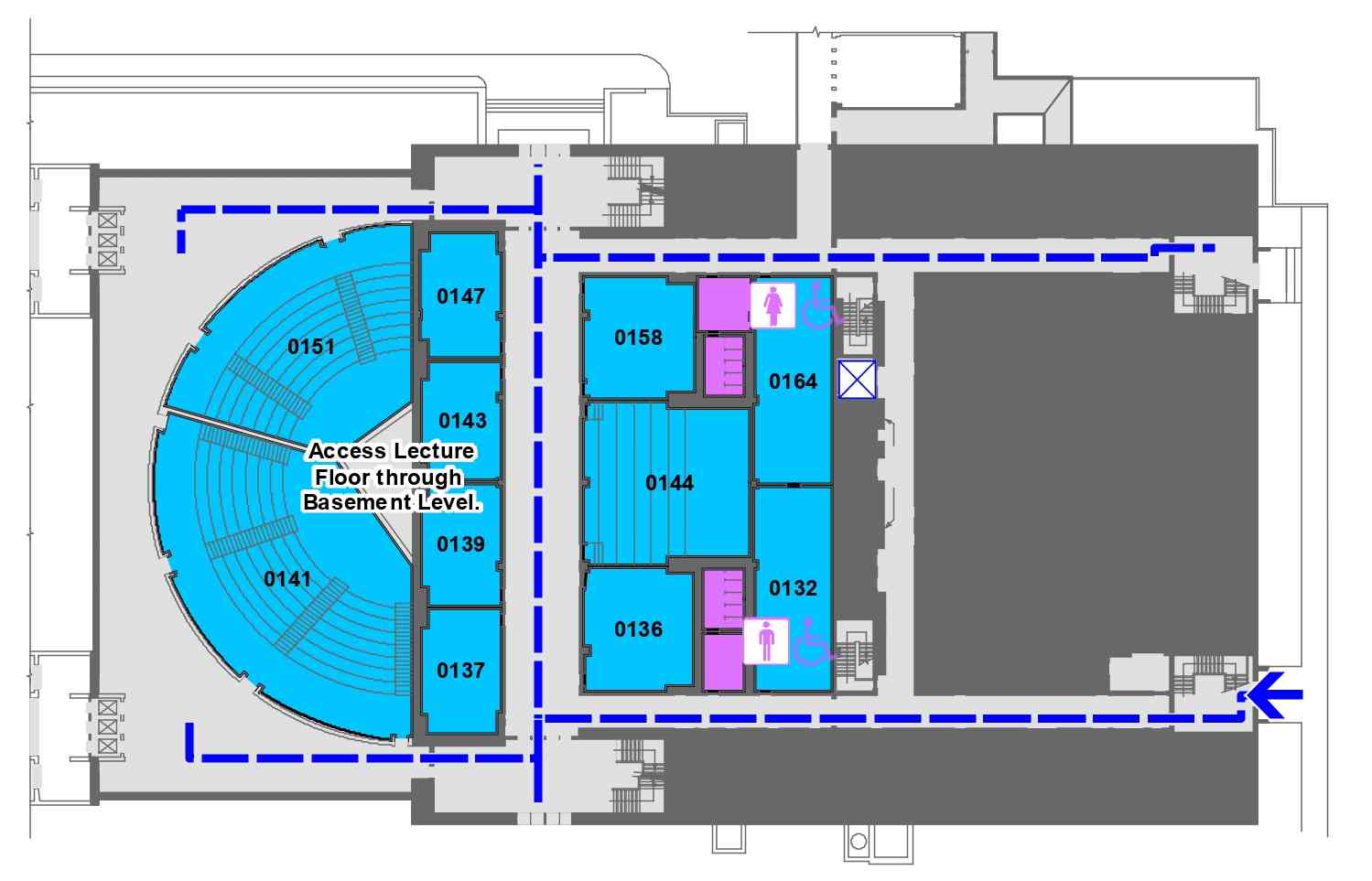}} 
\subfloat[UJI Library 5th Floor]{\includegraphics[width=0.65\columnwidth]{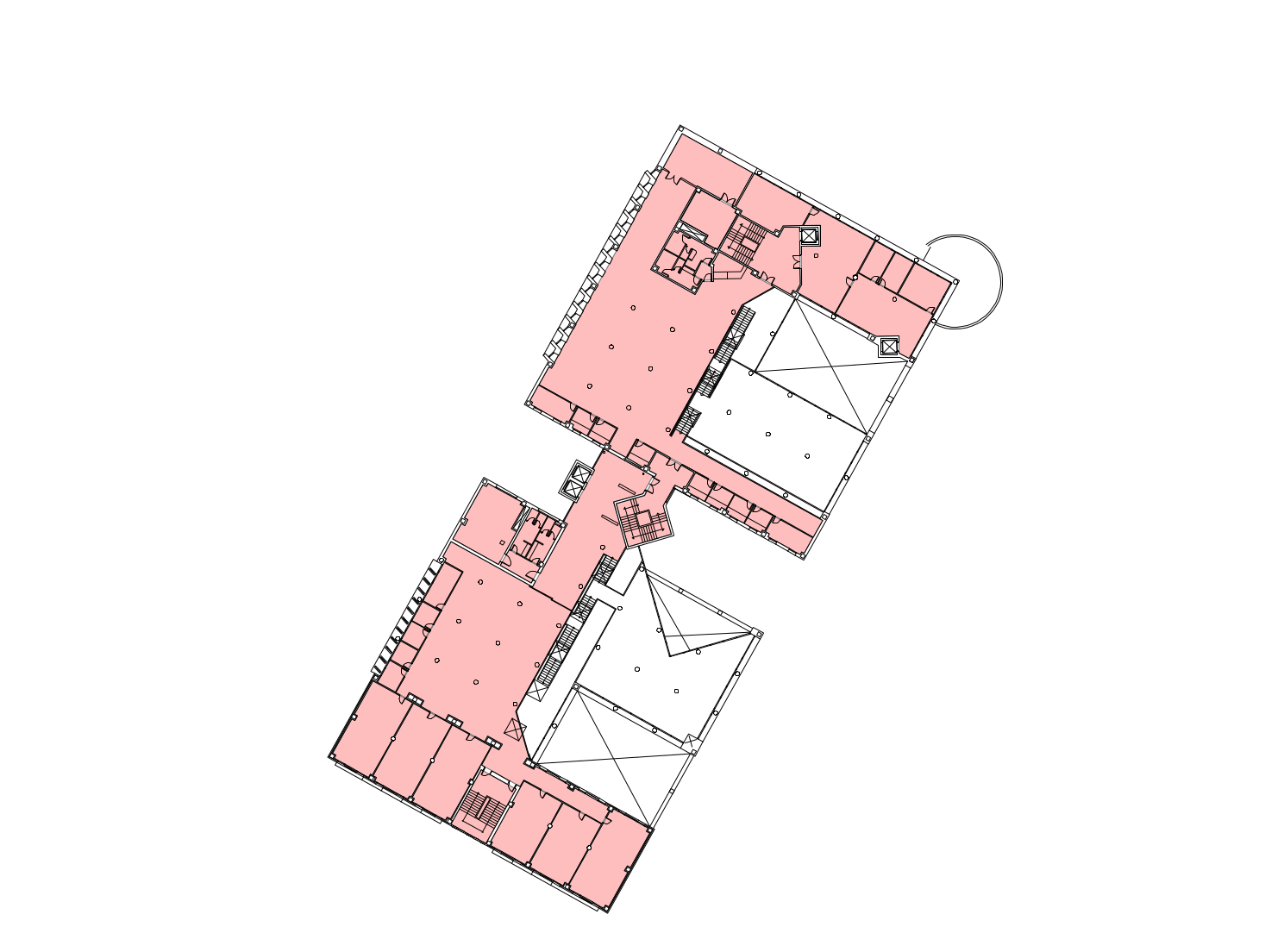}} 
\subfloat[Hyundai Department Store 4th Floor]{\includegraphics[width=0.65\columnwidth]{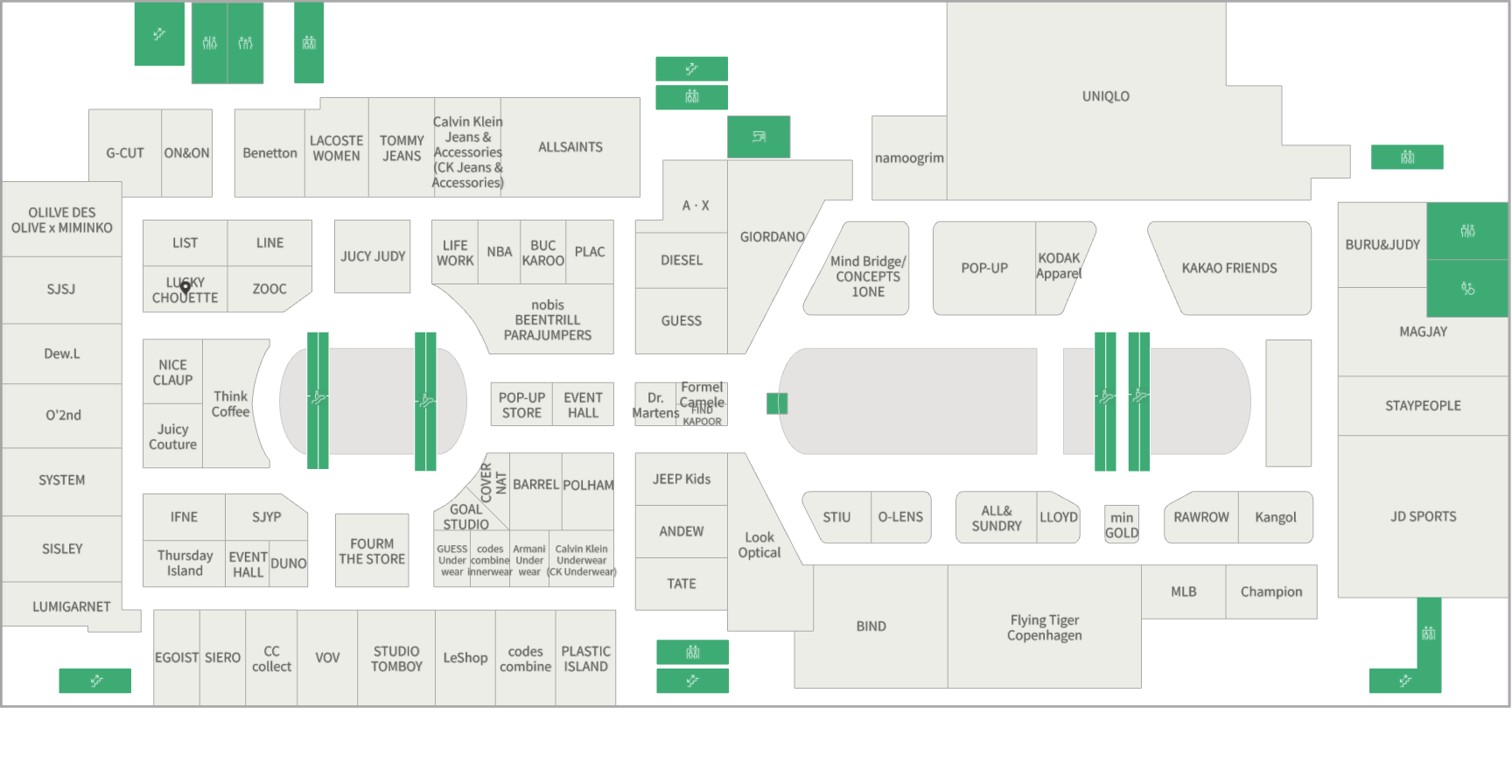}} 
\caption{Floor plans of the three different datasets that were used for evaluation.}
\label{fig:magpie_plans}
\end{figure*}

\vspace{-3mm}
\subsection{Datasets}
\label{ssec:datasets}

%
%
The {\it MagPIE dataset} was collected in three different buildings on the UIUC campus: the Coordinated Sciences Laboratory (CSL), Talbot Laboratory, and Loomis Laboratory~\cite{magpie_article}. 
The recorded trials cover all the buildings tightly. For each building, 3 tracks in the train set are retained as validation set. We use the validation tracks to evaluate the models and fine-tune the hyperparameters.


The dataset contains IMU (inertial measurement unit) and magnetometer measurements along with ground truth position measurements that have centimeter-level accuracy. The mobile phone magnetometer was calibrated by the authors of the dataset before starting a new trial.



\textit{IPIN'20 dataset.}
\label{ssec:ipin_dataset}
This dataset was collected as a part of the Track 3 IPIN'20 competition~\cite{potorti20ipin} in the five store library building located at Universitat Jaume I (Castell\'on, Spain). The dataset was collected by the same actor with a Samsung Galaxy A5 2017 (SM-A520F) smartphone with Android 8.0 using a dedicated ``GetSensorData'' app~\cite{ipin_data_collection_app}. For the experiments, we have selected $18$ short single-floor tracks (collected $4$ times each) and $4$ long trajectories across library bookshelves (two on floor 3 and two in floor 5). 
Out of these four tracks, two were chosen for train, one for validation and one for test. 

\textit{Hyundai Department store dataset.}
\label{ssec:hyundai_dataset}
This dataset was collected by two navigation robots, with smartphones being placed in special holders (see Figure~\ref{fig:photos}). The electromagnetic field generated by the robots 
heavily disturbs the analysis and modeling nearly magnetic anomalies. 

Moreover, the train and test sets were collected with four months of difference, by different robots.
Finally, during the train and test data collection runs the robot was not moving at exactly the same speed, and these engine  disturbances between train and test runs were also slightly different. 
Consequently, there is a severe mismatch of MF data recorded by four phones in the train and test sets, even at the same location. 
In Section~\ref{sec:aligment}, we addressed this mismatch problem and proposed an alignment of the collected magnetic data, before using it in our pipeline. 

\vspace{-5mm}
\subsection{MagPie evaluation results} 
\label{ssec:regression}
Table~\ref{tab:summary} presents evaluation results for three buildings in the MagPie dataset.
It compares the localization errors of our CNN+FN and CNN+RNN regression methods to the baseline CNN+FN Landmark-based classification method. 

The CNN+FN regression models show good results for CSL, but report high errors for other two buildings. 
The table includes the ablation analysis of the number of channels $N$. 
It reports localization errors when the number of channels is 1 ($x$ for RP), 3 ($x$, $y$, $x$ for RP), 9 ($x$, $y$, $z$ for RP, GASF and GADF) or 12 ($x$, $y$, $z$ for four 1D-to-2D transformation methods). Results clearly show that using multiple (alternative) channels greatly contribute to reducing the localization error.

Using landmarks for classification confirmed that buildings differ greatly in their magnetic anomalies. Good results are obtained only for CSL; in general, there is little significant improvement over CNN+FN regression.

Taking the trajectory context into account helps solve the pattern ambiguity problem and allows to reduce considerably the localization error.
In all models, we assume the starting point estimation is noisy for both training ans test trials. The noise is simulated as the normal distribution with the mean of 0 m  and variance of 3 m. 


\begin{table}[ht]
\centering
\begin{tabular}{ |c||c||c|c|c| }
\hline
Building & N      & CNN+FN&CNN+FN& CNN+RNN \\ 
         &channels& Landmarks & Regression & Regression\\ \hline
CSL      &  1     & 5.15  & 5.09 & 5.80 \\ \cline{2-5}
         &  3     & 2.16  & 1.47 & 4.61 \\ \cline{2-5}
         &  9     & 1.16  & 0.97 & 0.81 \\ \cline{2-5}
         & 12     & 0.95  & 0.98 & {\bf 0.30} \\ \hline \hline
Loomis   & 1  & 8.13 &  8.50 &  7.36 \\ \cline{2-5}   
         & 3  & 6.62 &  6.72 &  2.51 \\ \cline{2-5} 
         & 9  & 5.77 &  6.16 &  1.15 \\ \cline{2-5} 
         & 12 & 4.62 &  5.05 &  {\bf 1.07} \\ \hline \hline
Talbot   & 1  & 9.27 &  11.32 &  6.91 \\ \cline{2-5}
         & 3  & 6.79 &  6.91 &  4.04 \\ \cline{2-5}
         & 9  & 4.95 &  4.90 &  1.17 \\ \cline{2-5}
         & 12 & 4.49 &  4.72 &  {\bf 1.06} \\ \hline
\end{tabular}
\caption{Localization error of the deep regression and landmark classification methods for 1, 3, 9 and 12 channels.}
\label{tab:summary}
\end{table}
\begin{table}[ht]
\centering
\begin{tabular}{ |c||c|c| }
\hline
Floor                &  CNN+FN       &   CNN+RNN \\ 
                     &  Regression   &   Regression\\ \hline
1                    &  0.76         &   0.42 \\ \hline
2                    &  0.68         &   0.50 \\ \hline
\hspace{0.1cm} $3^+$ &  1.55         &   1.38 \\ \hline
\hspace{0.1cm} $3^-$ &  0.79         &   0.59 \\ \hline
4                    &  0.50         &   0.39 \\ \hline
\hspace{0.1cm} $5^+$ &  2.58         &   2.20 \\ \hline
\hspace{0.1cm} $5^-$ &  1.89         &   1.09 \\ \hline 
\end{tabular}
\caption{Localization error of the deep regression using CNN and RNN models on IPIN'20 data using the best set of hyper-parameters for each floor. {+/-} indicate passages of that floor with or without books shelves.}
\label{tab:summary-ipin}
\end{table}
\begin{table}[ht]
\centering
\begin{tabular}{ |c||c|c| }
\hline
Alignment type           & CNN+FNN  &CNN+RNN \\ 
                  &  Regression &Regression\\ \hline
No alignment   &   47.35 & 35.12 \\ \hline
Linear         &   3.72 & 1.43 \\ \hline
Deep           &   2.15 & 1.08 \\ \hline

\end{tabular}
\caption{Localization error (in meters) of the deep regression using CNN and RNN models on Hyundai Department Store (Floor 4) data using the best set of hyper-parameters without and with two different types of alignment of the magnetic data in the pre-processing step. }
\label{tab:summary-hyundai}
\end{table}

\vspace{-3mm}
\subsection{IPIN evaluation results} 
\label{ssec:ipin_eval}
Evaluation results for IPIN'20 dataset are reported in  
Table~\ref{tab:summary-ipin}. For each of the five floors in the dataset, it compares the localization errors for CNN+FN and CNN+RNN regressions. Moreover, for 3rd and 5th floors, it includes results for two evaluation versions,  without (easier case) and with (harder case) passages between the book shelves. As the table shows, in this dataset CNN+RNN regression models allow for 11\% to 45\% error reduction w.r.t. CNN+FN regression ones.

\vspace{-3mm}
\subsection{Hyundai evaluation results} 
\label{ssec:hyundai_eval}
Table~\ref{tab:summary-hyundai} shows evaluation results for Hyundai store dataset. It reports localization errors for CNN+FN and CNN+RNN regression models and for the linear and deep alignment methods presented in Section~\ref{sec:aligment}. For the sake of completeness, we also present results for both methods without alignment.
Like in the previous datasets, CNN+RNN regression models outperform CNN+FN regression ones. Moreover, the deep alignment provides a comfortable advantage over the linear one. 
\section{Conclusion}
\label{sec:conclusion}

In this paper, we presented a unique pipeline for MF-based indoor localization for two different setups, when MF data is ``clean" or disturbed by robot electro-magnetic field. 
In the latter case, we complete the pipeline with 
the deep alignment step to compensate data mismatch. It has an impact on the training and evaluation protocol, but permits to keep the localization error around 1 m in both setups.
This makes MF data-based localization competitive with other sensor-based localization methods. However, unlike Wi-Fi or Bluetooth based localization that are infrastructure-dependant, MF data based localization requires no infrastructure investment and counts on the advanced modeling of magnetic field anomalies in indoor environment only.

The main components of out pipeline
extend the state-of-the-art techniques
based on the identification of landmarks in magnetic map.
We brought multiple improvements in the localization process, including converting magnetic field time series into 2D representation to enable training CNN- and RNN-based models, processing the localization as deep regression. 

We tested our methods on two public and one proprietary datasets. Evaluation results show that our methods outperform the baseline methods by large margins, and report the localization error around 1 m. in all tested cases and situations. We also discussed limitations of our approach. 

\paragraph*{Acknowledgement}
We would like to thank to our colleagues from Naver Labs Korea for collecting  invaluable data at the Hyundai Department Store and Yohann Cabon, Philippe Rerole and the 3D Vision Group Lead Martin Humenberger from Naver Labs Europe for converting this data into the Kapture format~\cite{kapture}.

{\small
\bibliographystyle{ieee_fullname}
\bibliography{biblio}

\begin{thebibliography}{10}\itemsep=-1pt

\bibitem{ashraf20sensors}
Imran Ashraf, Soojung Hur, and Yongwan Park.
\newblock Enhancing performance of magnetic field based indoor localization
  using magnetic patterns from multiple smartphones.
\newblock {\em Sensors}, 20(9):2704, 2020.

\bibitem{ashraf_floor_2019}
Imran Ashraf, Soojung Hur, Muhammad Shafiq, and Yongwan Park.
\newblock Floor {Identification} {Using} {Magnetic} {Field} {Data} with
  {Smartphone} {Sensors}.
\newblock {\em Sensors}, 19(11):2538, Jan. 2019.

\bibitem{ashraf20minloc}
Imran Ashraf, Mingyu Kang, Soojung Hur, and Yongwan Park.
\newblock Minloc: Magnetic field patterns-based indoor localization using
  convolutional neural networks.
\newblock {\em IEEE Access}, 8:66213--66227, 2020.

\bibitem{potorti20ipin}
Francesco~Potorti at al.
\newblock The {IPIN} 2019 indoor localisation competition - description and
  results.
\newblock {\em {IEEE} Access}, 8:206674--206718, 2020.

\bibitem{chidlovskii19}
Boris Chidlovskii and Leonid Antsfeld.
\newblock Semi-supervised variational autoencoder for wifi indoor localization.
\newblock In {\em {Proc. IPIN}}, pages 1--8, 2019.

\bibitem{chung11}
Jaewoo Chung, Matt Donahoe, Chris Schmandt, Ig{-}Jae Kim, Pedram Razavai, and
  Micaela Wiseman.
\newblock Indoor location sensing using geo-magnetism.
\newblock In {\em MobiSys}, pages 141--154. {ACM}, 2011.

\bibitem{gru_article}
Kyunghyun~Cho et al.
\newblock Learning phrase representations using {RNN} encoder-decoder for
  statistical machine translation.
\newblock In {\em Proceedings {EMNLP}}, pages 1724--1734, 2014.

\bibitem{yu21calibration}
XiangQian~Yu et al.
\newblock Calibration of {AC} vector magnetometer based on ellipsoid fitting.
\newblock {\em {IEEE} Trans. Instrum. Meas.}, 70:1--6, 2021.

\bibitem{kapture}
Naver~Labs Europe.
\newblock A unified data format for visual localization, structure from motion
  and more.
\newblock \url{https://europe.naverlabs.com/research/3d-vision/kapture/}.

\bibitem{gu_indoor_2019}
Fuqiang Gu, Xuke Hu, Milad Ramezani, Debaditya Acharya, Kourosh Khoshelham,
  Shahrokh Valaee, and Jianga Shang.
\newblock Indoor localization improved by spatial context — a survey.
\newblock {\em ACM Comput. Surv.}, 52(3):64:1--64:35, July 2019.

\bibitem{magpie_article}
David Hanley, Alexander~B. Faustino, Scott~D. Zelman, David~A. Degenhardt, and
  Timothy Bretl.
\newblock Magpie: A dataset for indoor positioning with magnetic anomalies.
\newblock {\em {Proc. IPIN}}, 2017.

\bibitem{hirata21reccurence}
Yoshito Hirata.
\newblock Recurrence plots for characterizing random dynamical systems.
\newblock {\em Commun. Nonlinear Sci. Numer. Simul.}, 94:105552, 2021.

\bibitem{ipin_data_collection_app}
A.~R. Jim{\'e}nez, F. Granja, and J. Torres-Sospedra.
\newblock Tools for smartphone multi-sensor data registration and gt mapping
  for positioning applications.
\newblock {\em {Proc. IPIN}}, pages 1--8, 2019.

\bibitem{khoshro15}
Samaneh Khoshrou, Jaime~S. Cardoso, and Lu{\'{\i}}s~Filipe Teixeira.
\newblock Learning from evolving video streams in a multi-camera scenario.
\newblock {\em Machine Learning}, 100:609--633, 2015.

\bibitem{amid_article}
Namkyoung Lee, Sumin Ahn, and Dongsoo Han.
\newblock Amid: Accurate magnetic indoor localization using deep learning.
\newblock {\em Sensors}, 2018.

\bibitem{heading_estimation}
Thibaud Michel, Pierre Genev{\`e}s, Hassen Fourati, and Nabil Laya{\"i}da.
\newblock {Attitude Estimation for Indoor Navigation and Augmented Reality with
  Smartphones}.
\newblock {\em {Pervasive and Mobile Computing}}, 46:96--121, Mar. 2018.

\bibitem{surveyRP2020}
Tuna~D. Pham.
\newblock {\em {Fuzzy} {Recurrence} {Plots} and {Networks} with {Applications}
  in {Biomedicine}}.
\newblock Springer, 2020.

\bibitem{cyclic_lr}
Leslie~N. Smith.
\newblock Cyclical learning rates for training neural networks.
\newblock In {\em {Proc.} {WACV}}, pages 464--472, 2017.

\bibitem{song14}
Junyeol Song, Hyunkyo Jeong, Soojung Hur, and Yongwan Park.
\newblock Improved indoor position estimation algorithm based on geo-magnetism
  intensity.
\newblock In {\em {Proc.} {IPIN}}, pages 741--744. {IEEE}, 2014.

\bibitem{kalyan_16_locateme}
Kalyan~Pathapati Subbu, Brandon Gozick, and Ram Dantu.
\newblock Locateme: Magnetic-fields-based indoor localization using
  smartphones.
\newblock {\em {ACM TIST}}, 4(4), 2013.

\bibitem{gaf_mtf_article}
Zhiguang Wang and Tim Oates.
\newblock Imaging time-series to improve classification and imputation.
\newblock {\em Proc. IJCAI}, 2015.

\bibitem{yin17nlp}
Wenpeng Yin, Katharina Kann, Mo Yu, and Hinrich Sch{\"{u}}tze.
\newblock Comparative study of {CNN} and {RNN} for natural language processing.
\newblock {\em CoRR}, abs/1702.01923, 2017.

\bibitem{zafari2017a}
Faheem Zafari, Athanasios Gkelias, and Kin Leung.
\newblock A survey of indoor localization systems and technologies.
\newblock {\em CoRR}, abs/1709.01015, 2017.

\end{thebibliography}
}

\end{document}